\documentclass[conference]{IEEEtran}
\IEEEoverridecommandlockouts

\usepackage{cite}
\usepackage{multirow}
\usepackage{amsmath,amssymb,amsfonts}
\usepackage{algorithmic}
\usepackage{graphicx}
\usepackage{textcomp}
\usepackage{subfigure}
\usepackage{xcolor}
\usepackage{bbding}
\usepackage[numbers,sort&compress]{natbib}
\def\BibTeX{{\rm B\kern-.05em{\sc i\kern-.025em b}\kern-.08em
    T\kern-.1667em\lower.7ex\hbox{E}\kern-.125emX}}
\begin{document}

\title{Self-Supervised Scene Flow Estimation with Point-Voxel Fusion and Surface Representation

\thanks{This work was supported in part by the National Natural Science Foundation of China under Grant 62271160 and 62176068, in part by the Natural Science Foundation of Heilongjiang Province of China under Grant LH2021F011, in part by the Fundamental Research Funds for the Central Universities of China under Grant 3072024LJ0803, in part by the Natural Science Foundation of Guangdong Province of China under Grant 2022A1515011527.}}

\author{
\IEEEauthorblockN{1\textsuperscript{st} Xuezhi Xiang}
\IEEEauthorblockA{\textit{Harbin Engineering University} \\
Harbin, China \\
xiangxuezhi@hrbeu.edu.cn}
\and
\IEEEauthorblockN{2\textsuperscript{nd} Xi Wang}
\IEEEauthorblockA{\textit{Harbin Engineering University} \\
Harbin, China \\
xwayuzu@163.com}
\and
\IEEEauthorblockN{3\textsuperscript{rd} Lei Zhang}
\IEEEauthorblockA{\textit{Guangdong University of Petrochemical Technology} \\
Maoming, China \\
zhanglei@gdupt.edu.cn}
\and
\IEEEauthorblockN{4\textsuperscript{th} Denis Ombati}
\IEEEauthorblockA{\textit{Harbin Engineering University} \\
Harbin, China \\
deniso2009@gmail.com}
\and
\IEEEauthorblockN{5\textsuperscript{th} Himaloy Himu}
\IEEEauthorblockA{\textit{Harbin Engineering University} \\
Harbin, China \\
himaloy@hrbeu.edu.cn}
\and
\IEEEauthorblockN{6\textsuperscript{th} Xiantong Zhen}
\IEEEauthorblockA{\textit{Guangdong University of Petrochemical Technology} \\
Maoming, China \\
zhenxt@gmail.com}
}

\maketitle

\begin{abstract}
Scene flow estimation aims to generate the 3D motion field of points between two consecutive frames of point clouds, which has wide applications in various fields. Existing point-based methods ignore the irregularity of point clouds and have difficulty capturing long-range dependencies due to the inefficiency of point-level computation. Voxel-based methods suffer from the loss of detail information. In this paper, we propose a point-voxel fusion method, where we utilize a voxel branch based on sparse grid attention and the shifted window strategy to capture long-range dependencies and a point branch to capture fine-grained features to compensate for the information loss in the voxel branch. In addition, since xyz coordinates are difficult to describe the geometric structure of complex 3D objects in the scene, we explicitly encode the local surface information of the point cloud through the umbrella surface feature extraction (USFE) module. We verify the effectiveness of our method by conducting experiments on the Flyingthings3D and KITTI datasets. Our method outperforms all other self-supervised methods and achieves highly competitive results compared to fully supervised methods. We achieve improvements in all metrics, especially EPE, which is reduced by 8.51\% on the KITTI$_o$ dataset and 10.52\% on the KITTI$_s$ dataset, respectively.
\end{abstract}


\section{Introduction}
Recent advances in autonomous driving and robotic intelligence has stimulated researcher's interest for scene flow, which has been widely studied for predicting the motion field between two frames of point clouds. 2D optical flow estimation generates image flow fields by calculating the instantaneous velocity vector features of pixels between frame sequences. Considering the limited 2D scene information, scene flow estimation utilizes point clouds to construct 3D point-level flow fields, which generate more refined local motion information and relative position relationships.

Most existing scene flow estimation methods are point-based methods\cite{b1,b2,b3,b18} which utilize PointNet\cite{b4} and its variants\cite{b5,b6} to extract features directly from the original point clouds, which contain rich fine-grained information. However, point-based methods are computationally expensive, and it is difficult to effectively capture long-distance dependencies by aggregating local neighborhoods of points through KNN. In addition, the accuracy of scene flow suffers from the disorder and density non-uniformity of point clouds.
To tackle these problems, some methods convert points into voxels\cite{b7} to encode coarse-grained information. Nevertheless, the loss of detail information reduces the accuracy of scene flow estimation.
Therefore, some existing methods\cite{b8,b9} simply combine these two strategies, utilizing point branch to extract precise spatial location information and voxel branch to capture long-range correlations.
However, these fusion methods ignore the mutual guidance of the two branches in the feature extraction process, and fail to effectively utilize features at different scales.
In addition, the large amounts of empty voxels weakens the network's attention to spatial geometric information greatly.
PVT\cite{b10} proposes a Transformer-based point-voxel fusion architecture, whose voxel branch adopts sparse window attention for the empty voxel problem, we borrow this idea in our proposed method and utilize the shift window strategy to avoid high computational complexity. Therefore, we propose a point-voxel fusion architecture, in which the point branch utilizes pointnet++ to extract fine-grained features, and the voxel branch uses sparse grid attention(SGA) and the shift window strategy\cite{b11} to capture long-range dependencies.
In the fusion process, different from existing methods\cite{b8,b9}, we take the point-voxel fusion features of each layer as the input of the next layer, and combine features of different scales through skip connections. 
By fusing the complementary information extracted from the two branches we obtain features which are more favorable for calculating the correlation between two frames of point clouds.
\begin{figure*}[!ht]
	\centerline{\includegraphics[width=\textwidth]{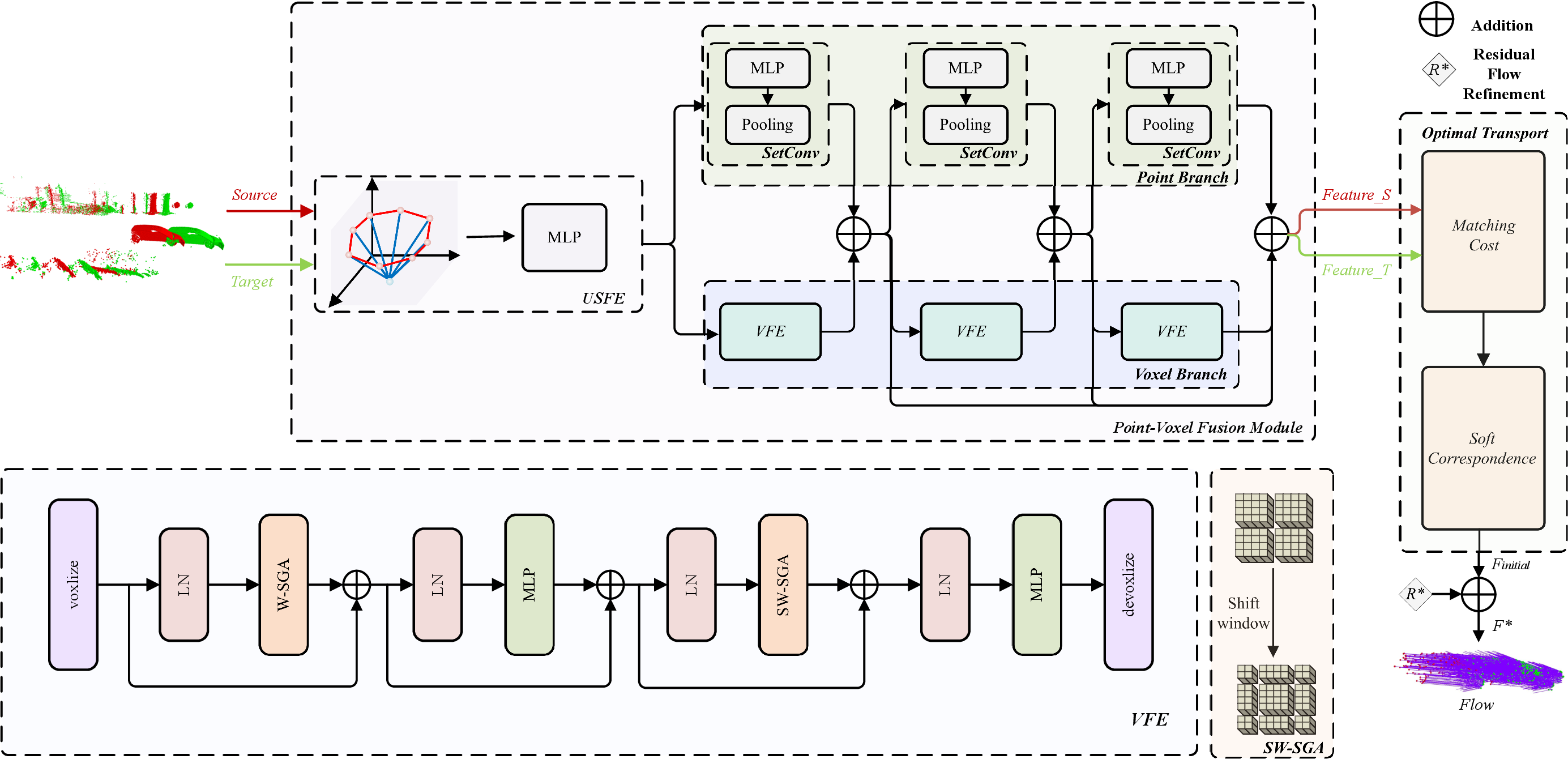}}
	\caption{Overview of our point-voxel fusion scene flow estimation network. The source frame and the target frame are input into the point-voxel fusion module to extract the deep features of the two frames of point clouds, respectively, and the weights are shared.}
\end{figure*}

Accurately extracting the surface information of 3D objects is essential to preserve the geometric structure of objects during the scene flow estimation process. Since various surfaces of objects exist various orientations in 3D space, it is difficult to fully represent the geometric structure by 3D coordinates. RepSurf\cite{b12} utilizes triangle and umbrella surfaces to depict the very local structure. SCF-Net\cite{b13} introduces a local polar representation block to construct a spatial representation which is invariant to the z-axis rotation. Inspired by RepSurf and SCF-Net, we utilize the Umbrella Surface Feature Extraction module to explicitly encode the local geometric features of the point cloud which is applied to both the point branch and the voxel branch. Our approach utilizes SCOOP\cite{b18} as the
baseline. Our contributions can be summarized as follows:
\begin{itemize}
\item We propose a point-voxel fusion method which effectively combines the fine-grained features extracted by the point branch and the coarse-grained features extracted by the voxel branch to capture local and long-range dependencies.
\item We utilize the USFE module to explicitly describe the surface structure of 3D objects which is combined with the point-voxel fusion architecture to make the features directionally sensitive.
\item Our network shows highly competitive performance on the FlyingThings3D and KITTI datasets.
Compared with the baseline, our EPE is decreased by 8.51\% on the KITTI$_o$ dataset and 10.52\% on the KITTI$_s$ dataset, respectively.
\end{itemize}

\section{METHOD}
Fig. 1 shows the overall framework of our network.
Given a source point cloud $S=\{p_i\in \mathbb{R} ^3\}_{i=1} ^N $ and a target point cloud $T=\{q_j\in \mathbb{R} ^3\}_{j=1} ^N $, our objective is to estimate the scene flow $F\in R^{N\times 3} $. 
First, we encode the local geometric features of the point cloud through USFE. Then the obtained geometric features and the 3D coordinates are fed into the point-voxel fusion architecture for deep feature embedding. After that, we utilize the features of the source frame and the target frame to compute the matching cost.
Following SCOOP, the optimal transport problem is solved based on the cost to compute the soft corresponding point for each source point, which are then subtracted from the source points to acquire the initial scene flow. Finally, we obtain the estimated scene flow through flow refinement.

\subsection{Umbrella Surface Feature Extraction}
The extraction of local geometric features is crucial for accurate scene flow estimation. In order to expand the perception field and obtain more stable local representations, for a point $p_i$, we sample $K$ neighborhood points $ \{p_i^1,p_i^2,...,p_i^k,...,p_i^K\}$ through KNN, where each neighborhood point is represented by Cartesian coordinates$(x_i^k,y_i^k,z_i^k)$. 
According to Cartesian coordinates, we order the points counterclockwise in the xy-plane to form an umbrella consisting of $K$ neighboring points and a centroid point. The centroid point is connected with each neighborhood point to form $K$ direction vectors, as shown in Equ. 1,
\begin{equation}d_i^k = p_i^k - p_i,\end{equation}
where $d_i^k$ represents the $k$-th direction vector of point $p_i$.

After that, we calculate the normal vector between the two neighboring direction vectors.
In order to keep the $K$ normal orientations consistent, we compute the cross-product with the direction vectors to obtain the normal features, as shown in Equ. 2,
\begin{equation}v_i^k=d_i^k \times d_i^{k+1},\end{equation}
where $v_i^k\in \mathbb{R} ^{1\times 3}$ represents the $k$-th normal feature of point $p_i$.

The direction information of $p_i$ is obtained by averaging normals.
In order to supplement the lack of angle information description in Cartesian coordinates, we include the spherical position of $p_i^k$ in its position information. The polar coordinates $(r_i^k,\theta_i^k,\phi_i^k)$ are calculated through Equ. 3,

\begin{equation}
	\left\{
	\begin{aligned}
		r_i&=\sqrt{x_i^2+y_i^2+z_i^2},\\
		\theta_i &=arctan\frac{z_i}{\sqrt{x_i^2+y_i^2}},\\
		\phi_i &=arctan\frac{y_i}{x_i}.\\
	\end{aligned}
	\right.
\end{equation}

Finally, we concatenate the normal information with the polar coordinates and achieve the surface structure features through stacked MLPs.

\subsection{Point-Voxel Fusion Module}
Our point-voxel fusion architecture is composed of a point branch and a voxel branch. 
The point branch contains three layers of SetConv based on PointNet++\cite{b5} architecture, where each layer consists of a multi-layer perceptron, instance normalization and a leaky ReLU activation.
In the voxel branch, we introduce sparse grid attention to extract relevant features, while utilizing a shift-window strategy to capture long-range dependencies. 

The voxel feature extraction module is shown in Fig. 1. 
Considering the impact of various scales of point clouds, we normalize the 3D coordinates $\{p_i\}$ of the input point cloud before voxelization, as shown in Equ. 4,
\begin{equation}\mu_i=\frac{p_i-\bar{p_i}}{\|p_i\|_2},\end{equation}
where $\mu_i$ is the normalized 3D coordinates and $\bar{p_i}$ is the coordinate of the gravity center point. And we constrain $\{\mu_i\}$ to the range [0,1] to reduce the burden of network training.

The normalized point cloud is represented as $ \{(\mu_i,t_i)\}$, where $t_i$ is the input point features. And then, we convert the normalized point cloud into voxel grids with resolution $r$.
The voxel index of each point mapping is calculated as shown in Equ. 5,
\begin{equation}V_{id}(u,v,w)=floor(\widehat{x_i}\times r,\widehat{y_i}\times r,\widehat{z_i}\times r),\end{equation}
where $floor$ is the rounding function. Then we average all point features inside each voxel to produce the voxel feature, as shown in Equ. 6,
\begin{equation}V_c=\frac{\sum_{k = 1}^{N}\mathbb{I}[V_{idk}]\times t_{k,c}}{N_{u,v,w}},\end{equation}
where $V_c$ is the feature of each voxel grid, $N_{u,v,w} $ represents the number of points in a voxel grid, $t_{k,c}$ denotes the $c$-th channel feature corresponding to $\mu_i$, and $\mathbb{I}[\cdot]$ is the binary indicator which determines whether the normalized point $ (\mu_k,t_{k})$ belongs to the voxel grid in Boolean form.
\begin{figure}[ht]
	\centerline{\includegraphics[width=\columnwidth]{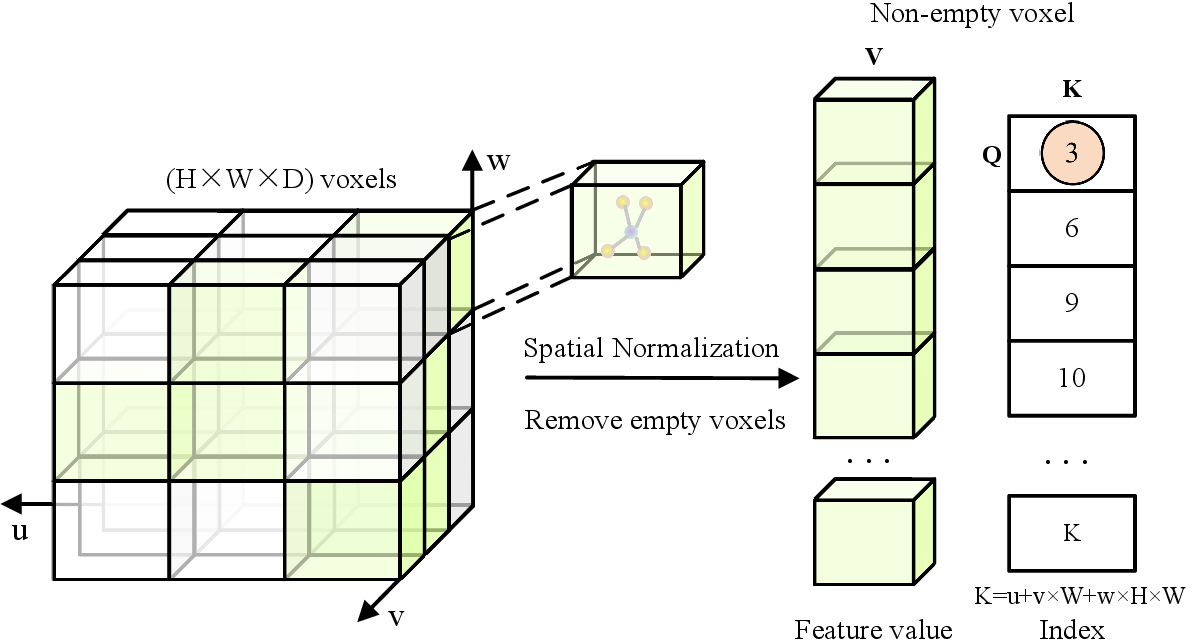}}
	\caption{Sparse Grid Attention. The values and coordinates of the non-empty voxels are stored in a 3D hash table, and then the coordinates are converted into index values as the Key in the attention calculation process.}
\end{figure}
After voxelization, we adopt a Transformer-based approach to extract voxel features. Due to the large number of empty voxels, directly performing self-attention on the entire voxel grid results in a waste of memory and high computational complexity. Therefore, the sparse grid attention is introduced to extract voxel features, as shown in Fig. 2.
First, the centroid of a non-empty voxel is assigned to the index of that voxel grid through a 3D hash table mapping. Then, a GPU-based rule book is utilized to store the voxel index and feature as the $Key$ and $Value$ respectively. After that, the coarse-grained features are obtained through sparse grid attention.
Considering the limited receptive field of sparse window attention, the shift window strategy is utilized to capture long-range contextual dependencies.

In order to combine long-distance dependencies and local features, we fuse the features extracted by the two branches. Before feature fusion, we utilize trilinear interpolation to project the voxel features into the point domain.

Finally we obtain the fusion features $F_{S}\in \mathbb{R} ^{N\times D}$ and $F_{T}\in \mathbb{R} ^{N\times D}$ of the source frame $S$ and the target frame $T$, as shown in Equ. 7 and Equ. 8,
\begin{equation}F_S=F_{point\_ S}+F_{voxel\_S},\end{equation}
\begin{equation}F_T=F_{point\_T}+F_{voxel\_T},\end{equation}
where $F_{point\_S}$ and $F_{voxel\_S}$ are the point features and voxel features of the source frame $S$, respectively. $F_{point\_T}$ and $F_{voxel\_T}$ are the point features and voxel features of the target frame $T$, respectively.

\section{EXPERIMENTS}
\begin{table}[h]
\caption{Performance comparison on KITTI$_o$ dataset. All methods are trained on FT3D$_o$ dataset. Bold indicates the best result. Underline indicates the second best result. Symbol $^+$ indicates that all points in the point cloud are used for evaluation\cite{b14,b15}.}
\centering
\renewcommand{\arraystretch}{1.2}
\setlength{\tabcolsep}{10pt}
\begin{tabular}{c|c|cccc} 
\hline
\multirow{1}{*}{Methods} &{Sup.} & \multicolumn{1}{c}{EPE$\downarrow$}& \multicolumn{1}{c}{AS$\uparrow$}& \multicolumn{1}{c}{AR$\uparrow$}& \multicolumn{1}{c}{Out$\downarrow$} \\ 
\hline
FlowNet3D\cite{b1}         & $Full$   & 0.173   & 27.6    & 60.9    & 64.9\\
FLOT\cite{b16}      & $Full$   & 0.107   & 45.1    & 74.0    & 46.3\\
BiPFN\cite{b3}        & $Full$   & 0.065   & 76.9    & 90.6   & 26.4\\
MSBRN\cite{b17}  & $Full$   & \underline{0.044}   & 87.3    & 95.0   & 20.8\\
\hline
SCOOP\cite{b18}         & $Self$   & 0.063      & 79.7       & 91.0    & 24.4\\
Ours          & $Self$   & 0.060   & 83.9    & 92.3    & 22.0\\
\hline
SCOOP$^+$\cite{b18}     & $Self$   & 0.047   & \underline{91.3}    & \underline{95.0}   & \underline{18.6}\\
Ours$^+$     & $Self$   & \textbf{0.043}   & \textbf{93.6}    & \textbf{95.4}   & \textbf{17.5}\\
\hline
\end{tabular}
\end{table}

\begin{table}[h]
\caption{Performance comparison on KITTI$_s$ dataset. All methods are trained on FT3D$_s$ dataset. Bold indicates the best result. Underline indicates the second best result.}
\centering
\renewcommand{\arraystretch}{1.2}
\setlength{\tabcolsep}{10pt}
\begin{tabular}{c|c|cccc} 
	\hline
	\multirow{1}{*}{Methods} &{Sup.} & \multicolumn{1}{c}{EPE$\downarrow$}& \multicolumn{1}{c}{AS$\uparrow$}& \multicolumn{1}{c}{AR$\uparrow$}& \multicolumn{1}{c}{Out$\downarrow$} \\ 
	\hline
	FlowNet3D\cite{b1}         & $Full$   & 0.177   & 37.4    & 66.8    & 52.7\\
	FLOT\cite{b16}          & $Full$   & 0.056   & 75.5    & 90.8    & 24.2\\
	PV-RAFT\cite{b8}         & $Full$   & 0.056  & 82.3    & 93.7   & 21.6\\
	DPV-RAFT\cite{b9} & $Full$   & 0.038   & 92.8    & 97.5   & 15.1\\
	MSBRN\cite{b17}  & $Full$   &\textbf{0.011}  & \underline{97.1}    & \textbf{98.9}   & \textbf{8.5}\\
	\hline
	PointPWC\cite{b2}   & $Self$   & 0.255   & 23.8    & 49.6   & 68.6\\
	SPFlow\cite{b20}   & $Self$   & 0.112   & 52.8    & 79.4   & 40.9\\
	FStep3D\cite{b21}      & $Self$   & 0.102   & 70.8    & 83.9   & 24.6\\
	SCOOP\cite{b18}  & $Self$   & 0.019      & 97.1       & 98.5    & 10.7\\
	Ours          & $Self$   & \underline{0.017}   & \textbf{97.2}    & \underline{98.6}    & \underline{10.5}\\
	\hline
\end{tabular}
\end{table}

\subsection{Implementation and Training Settings}
For FT3D$_o$ and KITTI$_o$ datasets, we evaluate our method on point clouds of 2048 points. And to completely evaluate the entire scene flow, we also utilize our method(Ours$^+$) to exploit the whole point cloud information and test the performance for the original resolution.
For FT3D$_o$ and KITTI$_o$ datasets, we use point clouds of 8192 points.

For the FT3D$_o$ dataset, we trained the model for 100 epoches with a batchsize of 4, and test on the KITTI$_o$ dataset. For the FT3D$_s$ dataset, we trained the model for 60 epoches with a batchsize of 1, and test on the KITTI$_s$ dataset. All our experiments were performed on a NVIDIA RTX 3060 GPU.

\subsection{Results}
We compare the results of training on the Flythings3D dataset and testing on the KITTI dataset with recent advanced works. TABLE I shows the results on FT3D$_o$ and KITTI$_o$ datasets. The results demonstrate that our method outperforms both self-supervised and fully supervised methods on all evaluation metrics. And we use only 1,800 randomly selected examples from FT3D$_o$, while the competitors employ all 18,000 scene instances.
Compared to the baseline, our approach shows a significant improvement in various metrics, which is because our point-voxel fusion architecture effectively combines the advantages of point features and voxel features, capturing detail information while obtaining long-range correlations.

TABLE II shows the results on FT3D$_s$ and KITTI$_s$ datasets. Our method provides higher $AS$ compared to MSBRN and yields competitive $EPE$, $AR$ and $Out$ results. Compared with the baseline SCOOP, our $EPE$ improves by 10.53$\%$. Moreover, our method has better results than PV-RAFT and DPV-RAFT, which also use point-voxel fusion architecture, demonstrating that our fusion method can extract features that are more conducive to computing the correlation between two frames of point clouds. Although the accuracy of our method is lower than that of the fully supervised method MSBRN, our model is learned in a self-supervised manner and achieves the best results among the self-supervised methods.

Fig. 2 shows a visualization comparisons on KITTI$_o$ dataset. We can see from the overall alignment of the target and predicted point cloud that our results outperform SCOOP.

\begin{table}[ht]
\caption{Results of Ablation Experiments}
\centering
\renewcommand{\arraystretch}{1.2}
\setlength{\tabcolsep}{10pt}
\begin{tabular}{ccccccccc}
	\hline
	USFE     & VFE    & EPE$\downarrow$    & AS$\uparrow$     &Params(M)  &FLOPs(G) \\ \hline
	&      & 0.063   & 79.7    &0.60  & 29.73   \\
	\Checkmark   &      & 0.062   & 83.5  &0.60  &31.10 \\
	\Checkmark  &\Checkmark    &\textbf{0.060}  &\textbf{83.9}  &{0.61} &50.14  \\
	\hline 
\end{tabular}
\end{table}

\subsection{Ablation Analysis}
In this section, we compare the impact of the introduction of different components, as shown in TABLE III. We first add the USFE to our baseline, with only point branch in the network. It can be seen that compared to the baseline, adding the USFE can improve the performance of scene flow estimation on all metrics, which proves that explicitly encoding local geometric structures is more conducive to preserving the edges and shapes of the 3D objects during the estimation process. Then we add the voxel branch to the baseline, and the network is a point-voxel fusion architecture. The surface features and 3D coordinates are fed into the branch together, which results in a 4.76$\%$ improvement in the $EPE$ metric compared to the baseline, demonstrating the effectiveness of our point-voxel fusion architecture. And the $Params$ only increases by 0.01M. Although the $FLOPs$ is increased, we achieve improvement in all metrics.

\begin{figure}
	\begin{center}
		\subfigure[SCOOP]{
			\includegraphics[width=1.6in,height=1in]{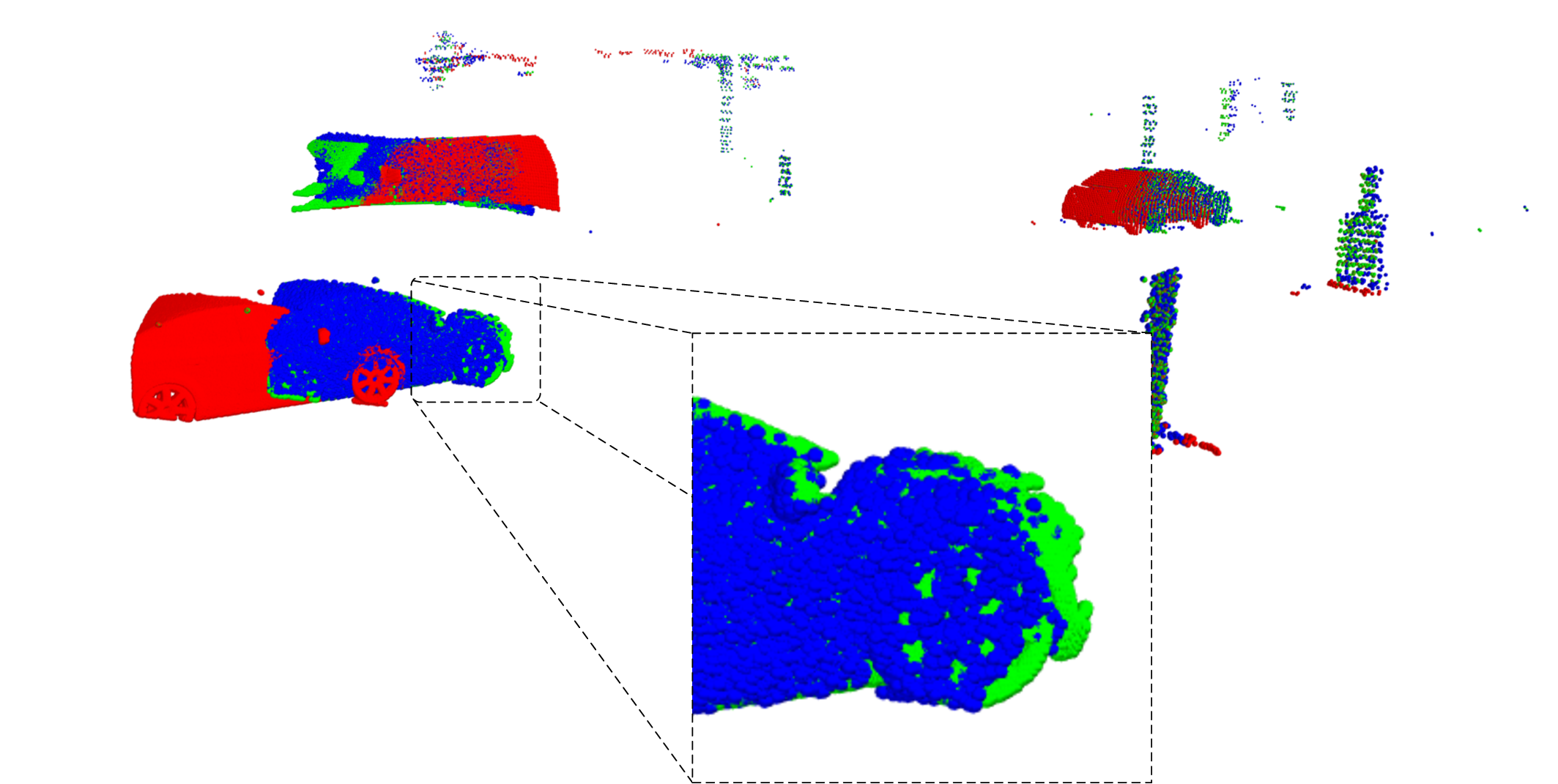}
		}
		\subfigure[Ours]{
			\includegraphics[width=1.6in,height=1in]{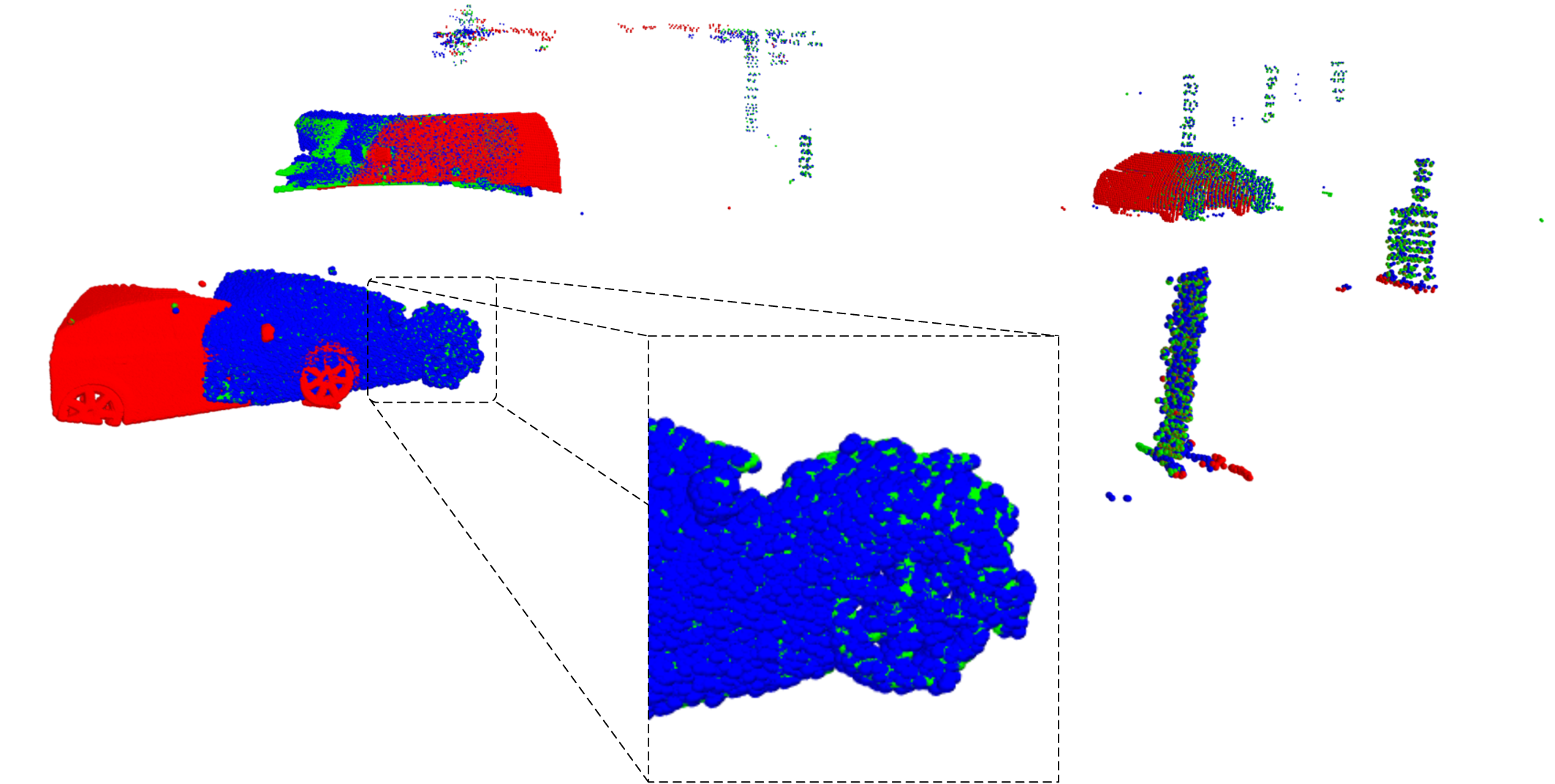}
		}
		\caption{Visual comparison on KITTI dataset.}
	\end{center}
\end{figure}

\section{Conclusion}
\label{conclusion}
In this paper, we proposed a scene flow estimation method based on point-voxel feature fusion. In order to estimate the similarity between point clouds more accurately, we fuse the point branch with the voxel branch to capture fine-grained features and long-range dependencies. And the USFE module utilizes explicit structural features of the point cloud to improve the accuracy of the network. We demonstrate the effectiveness of our proposed method by performing experiments with outstanding results on the FlyingThings3D and KITTI datasets.




\end{document}